# Design for Manufacturing: A Manufacturability Knowledge-Integrated Reinforcement Learning Framework for Free-Form Pipe Routing in Aeroengines

Caicheng Wang[1], Zili Wang[1,2*], Shuyou Zhang[1,2], Yongzhe Xiang [1], Zheyi Li[1], Liangyou Li[3], Jianrong Tan[1,2]

*(1. State Key Laboratory of Fluid Power and Mechatronic Systems, Zhejiang University, Hangzhou, 310027, China. 2. Engineering Research Center for Design Engineering and Digital Twin of Zhejiang Province, Zhejiang University, Hangzhou, 310027, China. 3. Zhejiang Changxing Heliang Intelligent Equipment Co., Ltd., Huzhou 313000, China))*

*E-mail address:* 12425135@zju.edu.cn *(Caicheng Wang),* ziliwang@zju.edu.cn *(Zili Wang), ,* zsy@zju.edu.cn *(Shuyou Zhang),* xyz2022@zju.edu.cn *(Yongzhe Xiang),* 12325137@zju.edu.cn *(Zheyi Li),* lily@zjheliang.com *(Liangyou Li),* eg2013@zju.edu.cn *(Jianrong Tan)*

**Corresponding author:* ziliwang@zju.edu.cn *(Zili Wang)*

**Abstract:**

Design for manufacturing plays a critical role in advanced aeroengine development, where complex components necessitate careful consideration of manufacturability. However, current practices in pipe routing remain largely decoupled from downstream manufacturing, leading to labor-intensive, trial-and-error iterations to achieve manufacturable designs. To address this problem, this study proposes the Frenet-based pipe routing optimization (FPRO) framework, a manufacturability knowledge-integrated reinforcement learning approach for free-form pipe design in aeroengines. FPRO formulates the routing problem as a boundary value problem in the Frenet frame. In this framework, the pipe path is represented by curvature and torsion profiles, which are generated using cubic Hermite interpolation. To integrate design and manufacturing, domain-specific manufacturing knowledge is embedded as constraints on the permissible ranges of curvature and torsion. The path optimization is performed using the proximal policy optimization algorithm with stochastic exploration and a stage-guided reward mechanism. A unified mapping formulation then translates the optimized path into motion trajectories for the bending die, enabling direct fabrication on a six-axis free-bending machine. Experimental results demonstrate that FPRO consistently generates collision-free, manufacturable paths with smoother geometric profiles compared to Cartesian-based methods. It also achieves faster convergence and superior performance in terminal alignment, path length, obstacle avoidance, and manufacturability compared to state-of-the-art reinforcement learning baselines. Real-world validation confirms the close geometric correspondence between the manufactured pipe and its digital design, validating the practical feasibility of FPRO. By integrating digital design and physical manufacturing, FPRO establishes an intelligent design-for-manufacturing paradigm that significantly reduces manual effort and design cycle time in aeroengine development.

## 1. Introduction

Design for manufacturing (DFM) is a discipline that focuses on incorporating constraints into the product design process to minimize manufacturing costs while ensuring the design meets functional requirements [1]. This concept is particularly critical in the aerospace industry, where the complexity of components necessitates careful consideration of manufacturability. As early as the 1960s, General Electric Corporation applied DFM principles and compiled comprehensive manufacturing data into a reference volume [2]. This resource was designed to provide designers with access to manufacturing knowledge, enabling them to create designs that can be manufactured efficiently. Despite the advantages of DFM, integrating complex manufacturing rules into the aeroengine design process remains a significant challenge, especially in pipe routing. The large number and diversity of pipes, along with the complexity of the pipe-bending process, significantly hinder coordination between design and manufacturing, often leading to schedule delays. To this day, the pipe design and manufacturing processes primarily involve design engineers developing conceptual designs, manufacturing engineers converting these concepts into manufacturable designs, and iterative redesigns leading to a final design that satisfies functional and manufacturability criteria [3]. Therefore, the gap between design and manufacturing in pipe routing still needs to be bridged further.

Pipe routing is a crucial subprocess of aeroengine design that focuses on establishing efficient connections between pipe inlets and outlets for transferring gases or liquid fuels within the confined three-dimensional space [4]. Typically, layout rules require the pipes to be collision-free and as short as possible to achieve lightweight and precise configurations. To achieve these objectives, researchers have developed effective automated pipe routing methodologies [5]. The pipe routing is modeled as a constrained path planning problem [6], and various algorithms have been employed to address this challenge. These include traditional point-to-point traversal or sampling methods [7], heuristic-based search algorithms [8], and more advanced, adaptive deep reinforcement learning (RL) algorithms [9]. Most layout research has focused on fixed-curvature pipes, particularly in shipbuilding and factory design [10, 11]. The growing structural complexity of aeroengines has recently driven the adoption of free-form pipes [12], which support more lightweight and sustainable design solutions.

Extensive research on pipe routing has enabled the rapid generation of layout solutions. Nevertheless, these solutions are often not directly applicable due to the omission of manufacturing constraints. On the one hand, various bending defects, such as cross-sectional distortion [13], axial springback [14], and wrinkling [15], inevitably occur during the bending process, causing deviations from the expected pipe shape. On the other hand, the structural and dimensional limitations of the bending machine constrain the range of manufacturable pipe specifications. These challenges are even more pronounced in the free-bending process, which typically restricts manufacturable geometries to pipes with large bending radii relative to their diameters [16]. Additionally, the free-bending process requires continuous curvature and torsion along the pipe path to prevent severe mechanical vibrations or impacts resulting from abrupt changes in die movement speed. Given these constraints, pipe routing in real-world engineering often requires close collaboration between design and manufacturing engineers, involving iterative design revisions to ensure that manufacturable solutions are achieved. This iterative process significantly increases design time and heavily relies on the engineers' experience, making it susceptible to human error.

Curvature and torsion of the pipe path are critical design parameters that govern the manufacturability of free-form pipes. Pipes with smooth curvature and torsion enhance the dynamic performance, stability, and bending accuracy of the bending machine [17]. Consequently, various smooth path planning methods can be applied in pipe routing to improve manufacturability by enhancing path smoothness. The most straightforward approach is to post-process the paths using spline curve fitting [18]. However, this method inevitably introduces deviations from the original planned path, potentially introducing new unknown collision risks. Several studies have focused on optimizing the control points of spline curves and incorporating curvature-related penalty terms into the optimization objective, ensuring smoothness is achieved inherently during the planning stage. For instance, smooth path planning for autonomous underwater vehicles has been achieved by combining the rapidly expanding random trees algorithm with quadratic B-spline curves [19]. This approach improves vehicle following performance by adjusting control point coordinates and limiting maximum curvature. To further enhance computational efficiency, another study employed Said-Ball curves and optimized control points through the slime mold algorithm, with the optimization objectives including maximum curvature, curvature derivative, and path length [20]. Discontinuities in path curvature and its derivatives can cause abrupt changes in velocity and acceleration, leading to slippage and over-actuation in mobile robots operating at high speeds [21]. To address this, higher-order Bézier curves [22] and simpler four-parameter logistic curves [23] have been employed for collision-free path planning with continuous curvature, ensuring higher-order continuity. Currently, most existing smooth path planning methods are designed for two-dimensional environments [24]. When extended to three dimensions, the planning complexity increases substantially due to the additional requirement of torsion control. Achieving $C^3$ continuity in 3D space requires at least a quartic B-spline curve, which significantly increases computational demands and the number of required initial conditions. Furthermore, in free-bending processes, the effects of curvature and torsion on the bending process are highly coupled and nonlinear, making it hard to ensure pipe manufacturability merely by incorporating penalty terms into the optimization objective.

Given these challenges, this study adopts a DFM paradigm for generating high-quality, reliable, and physically realizable pipe layouts. Specifically, we directly design the curvature and torsion profiles of the pipe and reconstruct the corresponding spatial path in a Cartesian frame using the Frenet–Serret formula [25]. In autonomous vehicle path planning, the Frenet frame is commonly employed to decouple vehicle motion in a Cartesian frame into lateral and longitudinal components relative to a road reference line, enhancing ride comfort, safety, and efficiency [26, 27]. In contrast, this study focuses on a three-dimensional scenario, aiming to design curvature and torsion functions along the pipe axis so that the reconstructed paths satisfy routing rules and manufacturability requirements.

This study formulates the free-form pipe routing within the Frenet frame as a boundary value problem (BVP). The curvature and torsion profiles of the pipe path are determined using cubic Hermite interpolation, and the selection of interpolation points is modeled as a sequential decision-making task solved by the proximal policy optimization (PPO) algorithm [28]. This RL-based algorithm is well-suited for sequential decision problems and exhibits greater adaptability to complex and uncertain three-dimensional environments than traditional path-planning or heuristic-based approaches [29, 30]. A fundamental challenge in RL is the exploitation-exploration dilemma, which involves balancing the selection of optimal actions to maximize rewards against exploring new states to obtain more information [31, 32]. To address this, we inject noise sampled from a uniform distribution into the mean of the action distribution, encouraging diverse actions. Furthermore, the complex routing task is decomposed into progressive sub-stages, with stage-guided rewards introduced to incentivize the agent's advancement through these stages, thereby fostering effective exploration. The main contributions of this research are summarized as follows:

1) To address the limited consideration of manufacturing constraints in prior pipe routing research, we analyze the free bending process to establish a unified mapping between pipe path geometry and die motion trajectories. A mathematical model of the manufacturing constraints is then formulated, providing an essential foundation for designing manufacturable pipe routes.

2) The manufacturability constraints are highly coupled and nonlinear in the Cartesian frame. To manage this complexity, we reformulate the pipe routing problem as a BVP in the Frenet frame. Curvature and torsion profiles are generated using cubic Hermite interpolation, and the corresponding spatial paths in the Cartesian frame are reconstructed using the Frenet–Serret formula. This transformation effectively decouples the influence of curvature and torsion on the bending process, facilitating compliance with manufacturability constraints.

3) The selection of interpolation points for curvature and torsion is modeled as a sequential decision-making problem, for which the robust PPO algorithm is employed as the core solution method. Stochastic noise is introduced into the action distribution to encourage exploration. The complex routing task is further decomposed into progressive sub-stages, with stage-guided rewards directing the agent through each phase. These strategies collectively enhance the reliability of action selection and improve the algorithm’s robustness in complex design environments.

The remainder of this study is organized as follows. Section 2 analyzes the free-bending process and derives a unified representation of die motion. Section 3 details the proposed Frenet-based pipe routing framework. Section 4 presents a comprehensive experimental validation, including comparisons with alternative methods and a demonstration of the framework's practical application. Finally, Section 5 concludes the study.

## 2. Motion analysis of free-bending process

Free-form pipes are typically manufactured using a six-axis free-bending machine, as illustrated in Fig. 1 (a). The free-bending process builds upon the principles of push bending, with the bending system comprising a pusher, a clamp die, a guider, and a bending die. During operation, the pipe is advanced by the pusher and passes sequentially through the clamp die, guider, and bending die. The pipe is then bent into a spatial shape through the coordinated motion of the bending die. The bending die has four degrees of freedom: translation along the *X* and *Y* axes and rotation about the *X* and *Y* axes, denoted as the *A* and *B* axes, respectively. The pusher provides a single degree of freedom of translation along the *Z*-axis, ensuring continuous pipe feeding. The *C*-axis, representing rotation about the *Z*-axis, is typically inactive during the bending process.

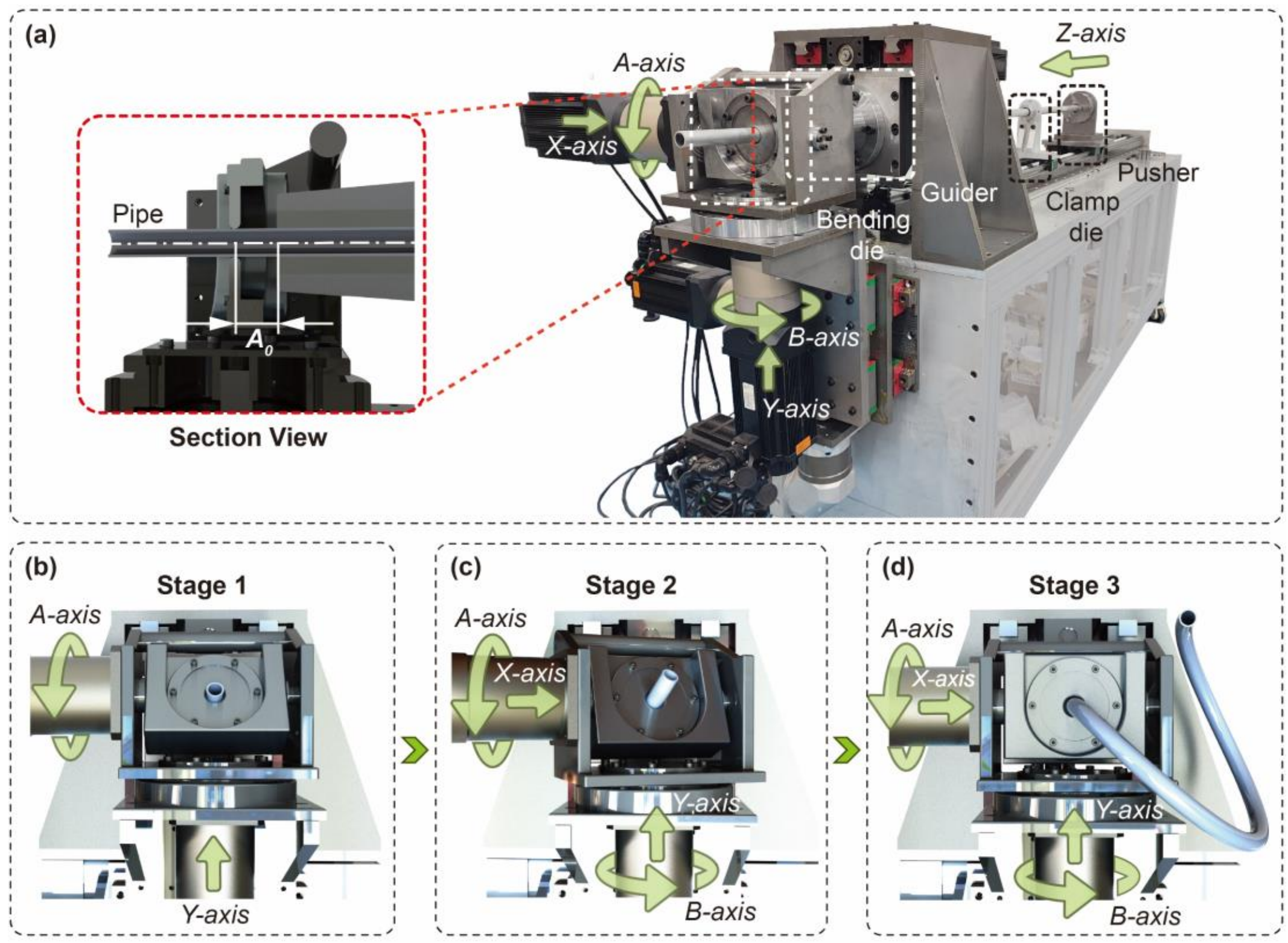

**Fig. 1.** Analysis of the free-bending process. (a) Six-axis free-bending machine. (b) Bending Stage 1. (c) Bending Stage 2. (d) Bending Stage 3.

First, consider the example of bending a cylindrical spiral pipe with a base radius $R_0$ and lead of $P_0$. Based on the motion characteristics, the free-bending process can be divided into three stages, as illustrated in Fig. 1 (b), (c), and (d).

- ***Stage 1:*** The bending die rotates clockwise by an angle $\alpha_A$ on the *A*-axis while translating a distance $U_y$ along the *Y*-axis. The parameters $\alpha_A$ and $U_y$ are given by:

$$\begin{cases} \alpha_A = \arcsin(\dfrac{kA_0}{R_0}) \\ U_y = R_0 - R_0\cos(\alpha_A) \end{cases} \tag{1}$$

where $k$ is an empirically defined correction factor, and $A_0$ represents the distance between the centers of the bending die and the guider shown in Fig. 1 (a). The pose of the bending die at the end of stage 1 can be expressed as:

$$.\begin{cases} P_x^1 = 0 \\ P_y^1 = U_y \\ \varphi_A^1 = -\alpha_A \\ \varphi_B^1 = 0 \end{cases}. \tag{2}$$

Stage 1 continues for a duration of $t_1$, defined as:

$$. t_1 = \frac{R_0 \alpha_A}{v_z} . \tag{3}$$

where $v_z$ is the speed of the pusher. This stage forms a transition section from the initial straight pipe to the spiral pipe.

- ***Stage 2:*** The bending die makes a rigid-body rotation around the *Z*-axis to form the main body of the spiral pipe. The total rotation angle is denoted as $\alpha_z$:

$$\alpha_z = \frac{2\pi l_0 P_0}{(2\pi R_0)^2 + P_0^2} \tag{4}$$

where $l_0$ denotes the arc length of the pipe. The pose of the bending die at the end of stage 2 can be expressed as:

$$\begin{cases} P_x^2 = U_y \sin(\alpha_z) \\ P_y^2 = U_y \cos(\alpha_z) \\ \varphi_A^2 = -\alpha_A \cos(\alpha_z) \\ \varphi_B^2 = \alpha_A \sin(\alpha_z) \end{cases} \tag{5}$$

Stage 2 lasts for a duration of $t_2$, defined as:

$$t_2 = \frac{l_0}{v_z} \tag{6}$$

- ***Stage 3:*** The bending die returns to its initial state, forming the transition section from the spiral pipe to the straight pipe.

The cylindrical spiral pipe represents the most fundamental spatial pipe that can be produced through the free-bending process owing to its constant curvature and torsion. As a result, the free-form pipe can be approximated as a sequence of connected spiral microelement segments with varying base radii and leads, thereby generalizing the bending process to cases with different curvatures and torsions. Consider sampling *n* points along the pipe's axis, where the curvature and torsion at the *i*th point are denoted as $k_i$ and $\tau_i$, respectively. The corresponding base radius and lead of the spiral microelement segment at the *i*th point can be derived as follows:

(7)

These sampling points divide the pipe into $n - 1$ spiral microelement segments. For the *i*th sampling point ($1 \le i \le n$), the parameters $\alpha_A$, $U_y$ and $\alpha_z$ can be computed as:

$$\begin{cases} \alpha_A^i = \arcsin(\frac{kA_0}{R_0^i}) \\ U_y^i = R_0^i - R_0^i \cos(\alpha_A^i) \\ \alpha_z^i = \frac{2\pi d^i P_0^i}{(2\pi R_0^i)^2 + (P_0^i)^2} \end{cases} \tag{8}$$

where $d^i$ denotes the distance between the $(i - 1)$th and *i*th sampling points, with $d^1$ defined as 0. Accordingly, the sequence of bending die poses can be expressed uniformly through the following formulation:

$$\begin{cases} P_x^i = U_y^i \sin(\sum_{j=1}^{i} \alpha_z^j) \\ P_y^i = U_y^i \cos(\sum_{j=1}^{i} \alpha_z^j) \\ \varphi_A^i = -\alpha_A^i \cos(\sum_{j=1}^{i} \alpha_z^j) \\ \varphi_B^i = \alpha_A^i \sin(\sum_{j=1}^{i} \alpha_z^j) \end{cases} \tag{9}$$

Equation (9) defines the pose of the bending die when the pipe reaches the *i*th sampling point. Notably, when $i = 1$, Eq. (9) simplifies to (2). Additionally, based on Eq. (3) and Eq. (6), the corresponding time series can be derived as follows:

$$t_i = \frac{R_0^1 \alpha_A^1}{v_z} + \sum_{i=1}^{n} \frac{d_i}{v_z} \tag{10}$$

Equations (7)–(10) provide a unified mapping from the curvature and torsion of a pipe to the corresponding motion trajectory of the bending die.

The manufacturability of free-form pipes is primarily influenced by two factors. The first is the structural size limitation of the dies, which requires $\frac{kA_0}{R_0^i} \leq 1$ to ensure the existence of $\alpha_A^i$. If there is a potential risk of interference between the bending die and the guide die, a more stringent upper bound on $\frac{kA_0}{R_0^i}$ may be imposed. The second factor is the material properties of the pipe, which impose a lower bound on the bending radius. A too-small bending radius may lead to defects such as cracking or excessive spring-back. Since both constraints impose a lower bound on $R_0^i$, they can be uniformly expressed as:

$$R_0^i = \frac{k_i}{k_i^2 + \tau_i^2} \geq R_{\min} \tag{11}$$

where $R_{\min}$ represents the minimum manufacturable bending radius. Furthermore, the curvature and torsion along the pipe axis must be continuous to ensure the smooth trajectory of the bending dies. Discontinuities in the trajectory can cause the die's instantaneous speed or acceleration to approach infinity, resulting in severe mechanical vibrations or impacts that compromise the system's dynamic stability and bending accuracy. These highly coupled and nonlinear constraints present key challenges in bridging the gap between pipe routing and practical manufacturing. In this study, the Frenet frame is innovatively employed to ensure that the generated path satisfies the bending radius constraints and continuity requirements.

## 3. Frenet-based pipe routing optimization framework

### *3.1. Frenet Frame-based problem modeling*

The Frenet Frame is a moving coordinate system in differential geometry used to describe the local properties of a smooth curve in three-dimensional space. As illustrated in Fig. 2 (a), let the vector function $\vec{r}(s)$ represent a point moving along the pipe axis in the Cartesian frame, where $s$ denotes the arc length parameter. The unit tangent, normal, and binormal vectors at $\vec{r}(s)$ are denoted by $\vec{T}(s)$, $\vec{N}(s)$, and $\vec{B}(s)$, respectively. Together, $\{\vec{T}(s), \vec{N}(s), \vec{B}(s)\}$ form a dynamic orthonormal coordinate system known as the Frenet frame (or TNB frame). The movement of the Frenet Frame along the curve is governed by the Frenet–Serret formula, expressed as:

$$\begin{cases} \dfrac{d\vec{T}(s)}{ds} = \kappa(s)\vec{N}(s) \\ \dfrac{d\vec{N}(s)}{ds} = -\kappa(s)\vec{T}(s) + \tau(s)\vec{B}(s) \\ \dfrac{d\vec{B}(s)}{ds} = -\tau(s)\vec{N}(s) \\ \dfrac{d\vec{r}(s)}{ds} = \vec{T}(s) \end{cases} \tag{12}$$

where $\kappa(s)$ and $\tau(s)$ denotes the curvature and torsion at $\vec{r}(s)$, respectively, assuming $\kappa(s) > 0$. Given $\kappa(s)$, $\tau(s)$, an initial position vector $\vec{r}_0$, and an initial frame orientation $\{\vec{T}_0, \vec{N}_0, \vec{B}_0\}$, Eq. (12) can be formulated as an initial value problem (IVP) with initial conditions as follows:

$$\begin{cases} \vec{T}(0) = \vec{T}_0 \\ \vec{N}(0) = \vec{N}_0 \\ \vec{B}(0) = \vec{B}_0 \\ \vec{r}(0) = \vec{r}_0 \end{cases} \tag{13}$$

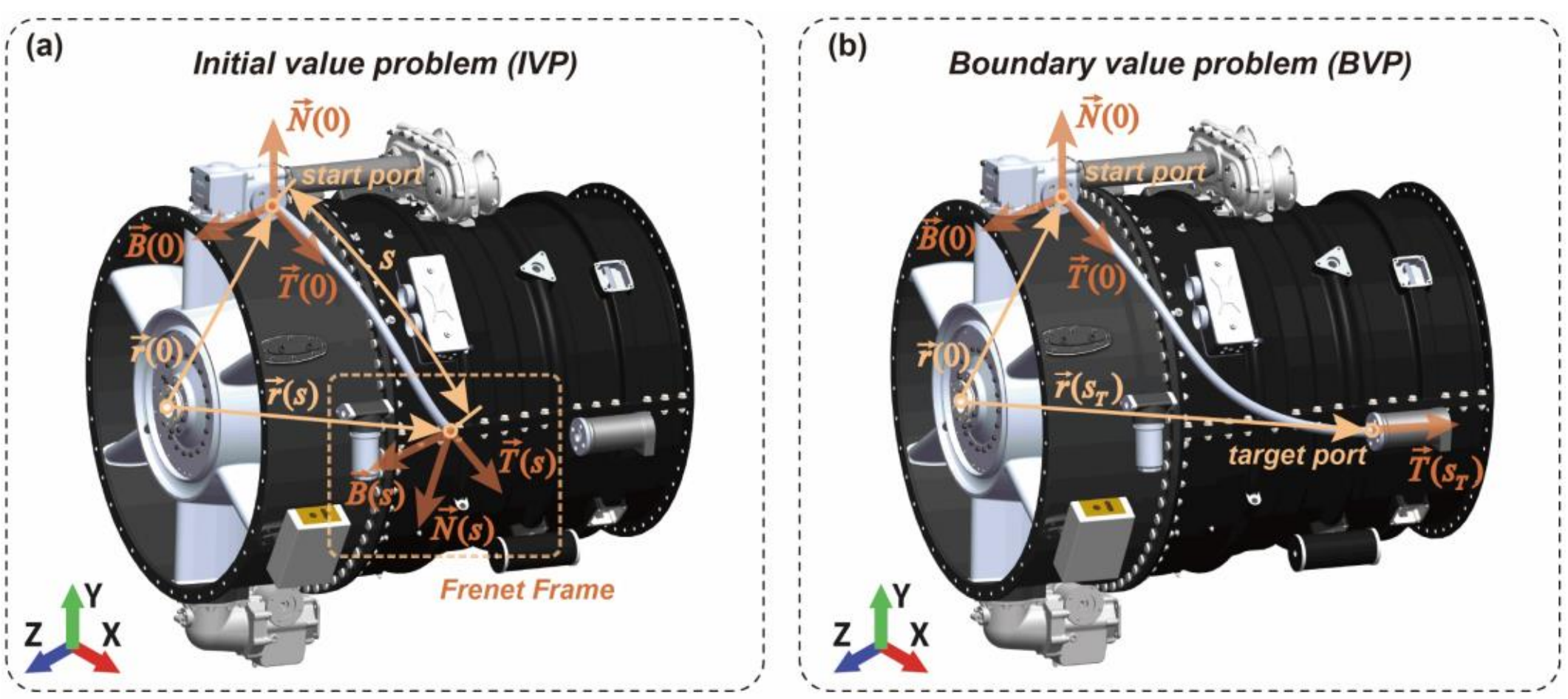

**Fig. 2.** Frenet Frame-based problem modeling. (a) Initial value problem. (b) Boundary value problem.

The trajectory of the solution in Eq. (12) can be numerically solved using integration techniques such as the Runge–Kutta method. As shown in Fig. 2 (b), the objective of the pipe routing task in aeroengine design is to plan a collision-free path that allows the pipe to extend smoothly from the start port to the target port. Consequently, in addition to the initial conditions, the vectors $\vec{T}(s)$ and $\vec{r}(s)$ must satisfy the following terminal conditions:

$$\begin{cases} \vec{T}(s_T) = \vec{T}_{\text{tar}} \\ \vec{r}(s_T) = \vec{r}_{\text{tar}} \end{cases} \tag{14}$$

where $s_T$ denotes the arc length at the end of the path, and $\vec{r}_{\text{tar}}$ and $\vec{T}_{\text{tar}}$ represent the specified target position and its corresponding tangent vector, respectively. Together, the initial and terminal conditions define the boundary conditions for Eq. (12), formulating a BVP.

The shooting method is a common technique for solving BVP. It reformulates the BVP as IVP, in which the initial guess parameters are iteratively adjusted until the solution satisfies the terminal boundary conditions. Specifically, the adjustable parameters of $\kappa(s)$ and $\tau(s)$ are continuously optimized to minimize the discrepancy between the computed terminal values and the expected target values $\vec{T}_{\text{tar}}$ and $\vec{r}_{\text{tar}}$, within a specified tolerance $\varepsilon$, such that:

$$\left\| \vec{T}(s_T) - \vec{T}_{\text{tar}} \right\| + \left\| \vec{r}(s_T) - \vec{r}_{\text{tar}} \right\| < \varepsilon \tag{15}$$

In addition to satisfying boundary conditions, pipe routing in this study must also ensure manufacturability, avoid obstacles, and minimize the overall pipe length. Considering all these constraints and optimization objectives, we formulated a composite loss function denoted as:

$$L_{total} = L_{BVP} + L_{obs} + L_{len} + L_{manuf} \tag{16}$$

where $L_{BVP}$ represents the loss of aligning the boundary conditions of BVP, $L_{obs}$ penalizes collision between the path and obstacles, $L_{len}$ penalizes path length, and $L_{manuf}$ penalizes violation of the manufacturability constraints defined in Eq. (11).

### *3.2. Cubic Hermite-based curvature and torsion construction*

The construction of $\kappa(s)$ and $\tau(s)$ is critical for solving the BVP, as these two functions fundamentally determine the solution trajectory. Their proper formulation directly affects whether the generated paths can satisfy the geometric and physical constraints of the pipe routing task. Specifically, $\kappa(s)$ and $\tau(s)$ must possess two key properties: sufficient flexibility to represent a wide range of spatial curves and compliance with the manufacturability constraints outlined in Section 2, including continuity requirement and minimum bending radius constraint.

To satisfy these requirements, we construct $\kappa(s)$ and $\tau(s)$ using piecewise cubic Hermite interpolation, as illustrated in Fig. 3. Taking $\kappa(s)$ as an example, suppose we have a set of interpolation points $\{(s_t, \kappa_t)\}_{t=0}^{T}$, with corresponding first-order derivatives $\{\kappa_t'\}_{t=0}^{T}$. For each interval $[s_t, s_{t+1}]$, a cubic polynomial can be constructed as follows:

$$H_t(s) = c_t^1(s - s_t)^3 + c_t^2(s - s_t)^2 + c_t^3(s - s_t) + c_t^4 \tag{17}$$

where the coefficients $c_t^1$, $c_t^2$, $c_t^3$, and $c_t^4$ are determined by the Hermite interpolation conditions:

$$\begin{cases} H_t(s_t) = \kappa_t \\ H_t(s_{t+1}) = \kappa_{t+1} \\ H_t'(s_t) = \kappa_t' \\ H_t'(s_{t+1}) = \kappa_{t+1}' \end{cases} \tag{18}$$

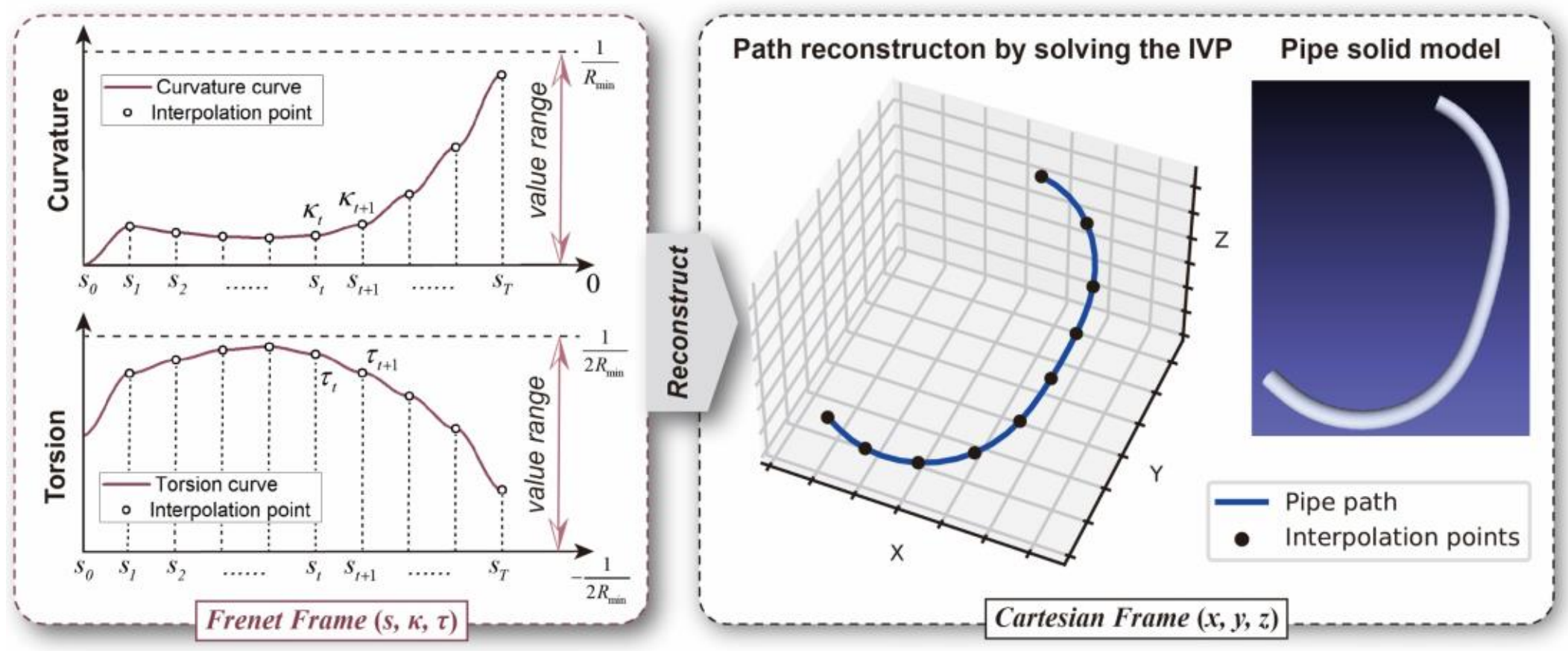

**Fig. 3.** Geometry properties construction using cubic Hermite interpolation and subsequent pipe path reconstruction.

These conditions ensure that the resulting curve has continuous values and first derivatives at each interpolation node. To further satisfy the minimum bending radius constraint, it is necessary to restrict the range of the interpolation points. Based on Eq. (11), the allowable ranges for $\kappa_t$ and $\tau_t$ are given by:

$$\begin{cases} -\dfrac{1}{2R_{\min}} \le \tau_t \le \dfrac{1}{2R_{\min}} \\ \dfrac{1 - \sqrt{1 - 4R_{\min}^2 \tau_t^2}}{2R_{\min}} \le \kappa_t \le \dfrac{1 + \sqrt{1 - 4R_{\min}^2 \tau_t^2}}{2R_{\min}} \end{cases} \tag{19}$$

It can be observed that the permissible range of $\kappa_t$ depends on the corresponding value of $\tau_t$. To decouple $\tau_t$ from $\kappa_t$, we relax the bounds on $\kappa_t$ by adopting the most conservative limits as follows:

$$0 \le \kappa_t \le \frac{1}{R_{\min}} \tag{20}$$

Although this relaxation may introduce the risk of violating the minimum bending radius constraint, such violations can be effectively penalized through the loss term $L_{manuf}$. Moreover, decoupling $\kappa_t$ and $\tau_t$ significantly benefits the subsequent optimization process. To further constrain the interpolation curve within acceptable limits, we set the first-order derivatives at each interpolation point to zero. Once the interpolation points are determined, the corresponding pipe paths in the Cartesian frame can be reconstructed by solving the IVP.

Indeed, various interpolation methods, such as linear or B-spline interpolation, can be used to construct $\kappa(s)$ and $\tau(s)$. However, piecewise cubic Hermite interpolation offers several key advantages. First, it is straightforward to implement, requiring no additional parameter tuning beyond specifying the interpolation points. Second, it exhibits strong locality, with each curve segment over $[s_t, s_{t+1}]$ depending only on the interpolation points at $s_t$ and $s_{t+1}$, thereby simplifying control and adjustment. Finally, it ensures continuity and sufficient smoothness to satisfy the manufacturability constraints. These properties make piecewise cubic Hermite interpolation a robust and efficient choice for defining the geometric profiles in our framework.

### 3.3. Interpolation points decision via stage-guided stochastic PPO

The selection of optimal interpolation points $\{(s_t, \kappa_t, \tau_t)\}_{t=0}^{T}$ is a high-dimensional optimization problem that can be framed as a sequential decision-making task. Traditional shooting methods, which rely on numerical optimization, are often ineffective in such complex search spaces. Therefore, this study adopts the robust PPO algorithm as the core of the shooting method, where an agent learns to determine the optimal points through iterative environmental interaction.

Within the RL framework, the task is represented as a Markov decision process, defined by $M = (O, A, R, p, \gamma)$, where $O$ denotes the observation space, $A$ represents the action space, $R$ is the reward function, $p$ indicates the state transition probability, and $\gamma$ is the reward discount factor. At each time step $t$, the agent receives an observation $o_t \in O$ from the environment, selects an action $a_t \in A$ based on the policy $\pi_\phi(a_t|o_t)$, and obtains a reward $r_t = R(o_t, a_t)$. The goal of the agent is to learn an optimal policy that maximizes the expected return $\mathbb{E}[\sum_{t=0}^{T} \gamma^t r_t]$ by optimizing the policy parameters $\phi$. The definitions of observation, action, and reward are detailed in the following sections.

#### 3.3.1. Observation space

To provide the agent with a comprehensive perception of the environment, the observation at time step $t$ is defined as $o_t = [o_c, o_f]$, where $o_c$ represents observations in the Cartesian frame, and $o_f$ contains information in the Frenet frame. Specifically, $o_c$ is defined as:

$$o_c = [x_t, y_t, z_t, \vec{d}_t] \tag{21}$$

where $(x_t, y_t, z_t)$ denote the Cartesian coordinates of the agent, and $\vec{d}_t$ is a vector representing distances from the agent to impassable regions in various spatial directions, providing local environmental context. The Frenet-frame-related observation $o_f$ is defined as:

$$o_f = [\kappa_t, \tau_t, \vec{T}(s_t), \vec{N}(s_t), \vec{T}_{\text{tar}} - \vec{T}(s_t), \vec{r}_{\text{tar}} - \vec{r}(s_t)] \tag{22}$$

where $\kappa_t$ and $\tau_t$ are the curvature and torsion at $s_t$, both initialized to zero when $t = 0$. $\vec{T}(s_t)$ and $\vec{N}(s_t)$ are the tangent and normal vectors at $s_t$, $\vec{T}_{\text{tar}} - \vec{T}(s_t)$ represents the deviation from the target tangent vector, and $\vec{r}_{\text{tar}} - \vec{r}(s_t)$ represents the deviation from the target position vector. Given the observation $o_t$, the agent selects an action $a_t$ according to the current policy $\pi_\phi(a_t|o_t)$.

#### 3.3.2. Action space with stochastic noise injection

The action $a_t$ at each time step $t$ is defined as:

$$a_t = [s_{t+1}, \kappa_{t+1}, \tau_{t+1}, \theta_t] \tag{23}$$

where $s_{t+1} \in (0, s_{max}]$ represents the arc length to the next interpolation point, while $\kappa_{t+1}$ and $\tau_{t+1}$ denote its curvature and torsion, respectively. The permissible ranges for $\kappa_{t+1}$ and $\tau_{t+1}$ are specified in Eq. (19) and Eq. (20). $\theta_t$ denotes the rotation angle of the Frenet frame around the tangent vector $\vec{T}(s_t)$. Notably, $\theta_t$ is only meaningful at $t = 0$, where it is used to adjust the initial orientation of the Frenet frame. For subsequent steps ($t > 0$), $\theta_t$ is fixed to zero.

In continuous action spaces, the PPO algorithm models the policy as a stochastic neural network that outputs a distribution over actions rather than deterministic values. Specifically, the policy network generates a mean vector $\mu_t$ and a logarithmic standard deviation vector $\log \sigma_t$ based on the current observation $o_t$. These parameters define a multivariate Gaussian distribution $N(\mu_t, \text{diag}(\sigma_t^2))$, from which the agent samples an action $a_t$ at each time step:

$$a_t \sim N(\mu_t, \text{diag}(\sigma_t^2)) \tag{24}$$

To enhance the exploration capabilities of the PPO agent in complex routing tasks, we adopt a method inspired by [33] that introduces stochastic noise to increase action entropy. Specifically, a uniform noise vector $z_t \sim U(-\alpha, \alpha)$, with the same dimensionality as $\mu_t$, is added to the policy mean to produce a perturbed mean $\mu_t' = \mu_t + z_t$. The action log probabilities required for computing the policy loss are then evaluated using the modified distribution $N(\mu_t', \text{diag}(\sigma_t^2))$. This additional stochasticity encourages the agent to explore diverse configurations within the action space, which is crucial for preventing convergence to suboptimal local solutions.

### *3.3.3. Stage-guided reward function*

In the pipe routing process, the agent must balance multiple, often conflicting, objectives. These include minimizing the total pipe length, avoiding collisions with obstacles, and adhering to manufacturability constraints. The inherent trade-offs among these objectives can lead to suboptimal outcomes, challenging the agent to decide between aggressive progression toward the target and cautious avoidance of constraint violations. Furthermore, sparse rewards and limited environmental perception may induce undesirable behaviors, such as greed, timidity, or recklessness.

To address these challenges, we decompose the complex routing task into four progressive sub-stages: startup, navigation, alignment, and shooting, as shown in Fig. 4. A stage-guided reward function is formulated to steer the agent sequentially through these stages. The total reward at time step *t*, denoted as $r_t$, is defined as the sum of an objective-driven reward $r_t^o$ and a stage bonus reward $r_t^s$:

$$r_t = r_t^o + r_t^s \tag{25}$$

Here, $r_t^o$ is a dense reward that provides continuous feedback based on the primary routing objectives and is active throughout all stages. In contrast, $r_t^s$ is a conditional reward activated only when specific conditions are met.

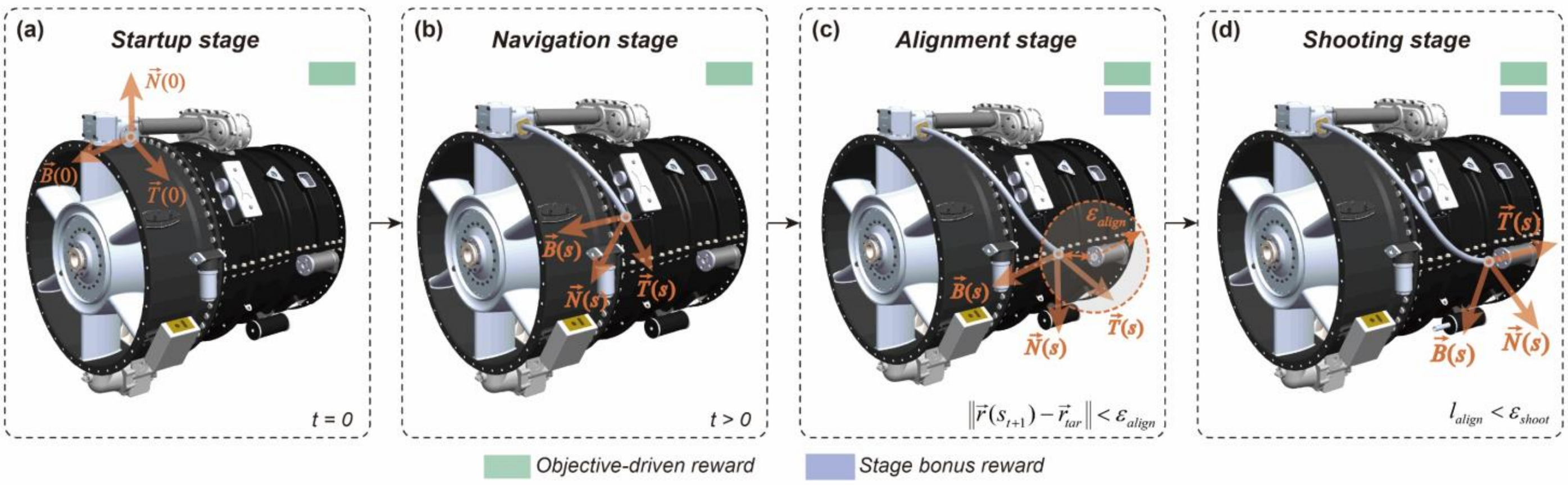

**Fig. 4.** Progressive sub-stages for the pipe routing process. (a) Startup stage. (b) Navigation stage. (c) Alignment stage. (d) Shooting stage.

The routing process begins in the startup stage (*t* = 0), where the agent determines an initial orientation by rotating the Frenet Frame at the starting point. Subsequently, the agent enters the navigation stage, advancing cautiously toward the target while avoiding collisions and constraint violations. During these two stages, the agent is guided exclusively by $r_t^o$, which is defined as a weighted sum of four components:

$$r_t^e = w_1 r_{dist} + w_2 r_{angle} + w_3 r_{len} + w_4 r_{obs} + w_5 r_{manuf} \tag{26}$$

where $r_{dist}$ encourages progress toward the target by rewarding reductions in the Euclidean distance between the agent's current position and the target port over consecutive steps. $r_{angle}$ incentivizes the agent to align its orientation with the target orientation by rewarding reductions in the angular difference. These terms are expressed as:

(27)

$$r_{angle} = \angle(\vec{T}(s_t), \vec{T}_{\text{tar}}) - \angle(\vec{T}(s_{t+1}), \vec{T}_{\text{tar}}) \tag{28}$$

where $\angle(\cdot,\cdot)$ denotes the angle between two vectors.

To favor shorter routes, $r_{len}$ is applied by penalizing the length of the newl $r_{dist} = \|\vec{r}(s_t) - \vec{r}_{\text{tar}}\| - \|\vec{r}(s_{t+1}) - \vec{r}_{\text{tar}}\|$ y generated pipe

segment:

$$r_{len} = -(s_{t+1} - s_t) \tag{29}$$

The terms $r_{obs}$ and $r_{manuf}$ represent penalties for obstacle collisions and violations of manufacturing constraints, respectively. To quantify these violations, a set of $N$ points $P = \{p_1, p_2, \dots p_N\}$ is sampled along the newly generated pipe segment. The penalties are then calculated as the fraction of sample points that are in violation:

$$r_{obs} = -\frac{1}{N}\sum_{i=1}^{N}\mathbb{I}_{obs}(p_i) \tag{30}$$

$$r_{manuf} = -\frac{1}{N}\sum_{i=1}^{N}\mathbb{I}_{manuf}(p_i) \tag{31}$$

where $\mathbb{I}_{obs}(p_i) = 1$ if $p_i$ intersects with an obstacle, and 0 otherwise. Similarly, $\mathbb{I}_{manuf} = 1$ if $p_i$ violates manufacturing constraints.

Once the agent's Euclidean distance to the target port falls below a predefined threshold $\varepsilon_{align}$, it transitions to the alignment stage. The objective here is to align the pipe's terminal segment with the target port direction. To quantify the alignment quality, we define an alignment loss $l_{align}$ as follows:

$$l_{align} = \frac{l_{angle} + l_{dist}}{2} \tag{32}$$

where $l_{angle}$ measures the normalized angular deviation and $l_{dist}$ measures the normalized perpendicular distance from the pipe endpoint $\vec{r}(s_t)$ to the line defined by the $\vec{r}_{tar}$ and $\vec{T}_{tar}$. These two components are defined as:

$$l_{angle} = \frac{1}{\pi}\arccos\left(\frac{\vec{T}(s_t)\cdot\vec{T}_{tar}}{\left\|\vec{T}(s_t)\right\|\left\|\vec{T}_{tar}\right\|}\right) \tag{33}$$

$$l_{dist} = \frac{\left\|\left(\vec{r}(s_t) - \vec{r}_{tar}\right)\times\vec{T}_{tar}\right\|}{s_{\max}\cdot\left\|\vec{T}_{tar}\right\|} \tag{34}$$

The stage bonus reward $r_t^s$ during this stage is defined as:

(35)

where $r_{bonus}$ is a fixed, one-time reward granted upon first entering the alignment or shooting stage, and $r_{align} = l_{align}(s_t) - l_{align}(s_{t+1})$ provides a continuous incentive for reducing the alignment loss, weighted by $w_{align}$.

When $l_{align}$ drops below a threshold $\varepsilon_{shoot}$, the agent enters the final shooting stage. The $r_t^s$ is updated to prioritize fine-grained alignment using a larger weight $w_{shoot}$:

$$r_t^s = r_{bonus} + w_{shoot} r_{align} \tag{36}$$

Once $l_{align}$ falls below a final tolerance $\varepsilon_{final}$, the routing task is deemed complete. The terminal boundary conditions of the BVP can then be met by extending a straight pipe segment to the target port.

The flowchart of our proposed framework, Frenet-based pipe routing optimization (FPRO), is illustrated in Fig. 5. The process begins in the design domain, with the specified start and target ports serving as initial inputs. In each iteration, the PPO agent observes the current environmental state and selects an action from its policy to determine the next interpolation point. This action is then used to construct a new segment for the curvature and torsion profiles over the interval $[s_t, s_{t+1}]$ via cubic Hermite interpolation. The corresponding pipe path segment in Cartesian space is reconstructed by solving the IVP associated with the Frenet-Serret formulas. Subsequently, our stage-guided reward function evaluates the newly added segment and provides feedback to update the PPO agent's policy parameters. Through this iterative loop of environment interaction and policy optimization, the agent progressively learns an optimal policy. This learned policy enables the generation of a complete pipe layout that satisfies the predefined boundary conditions while adhering to collision avoidance and manufacturability constraints.

A key advantage of FPRO is its seamless integration with the manufacturing domain. The final, optimized curvature and torsion functions from the design domain serve as direct inputs for the manufacturing process. Leveraging the unified mapping formula derived in Section 2, these geometric profiles are translated into the corresponding die motion trajectories for the six-axis free-bending machine. This framework establishes a direct and reliable transition from digital design to physical production, ensuring the high-quality fabrication of the final pipe.

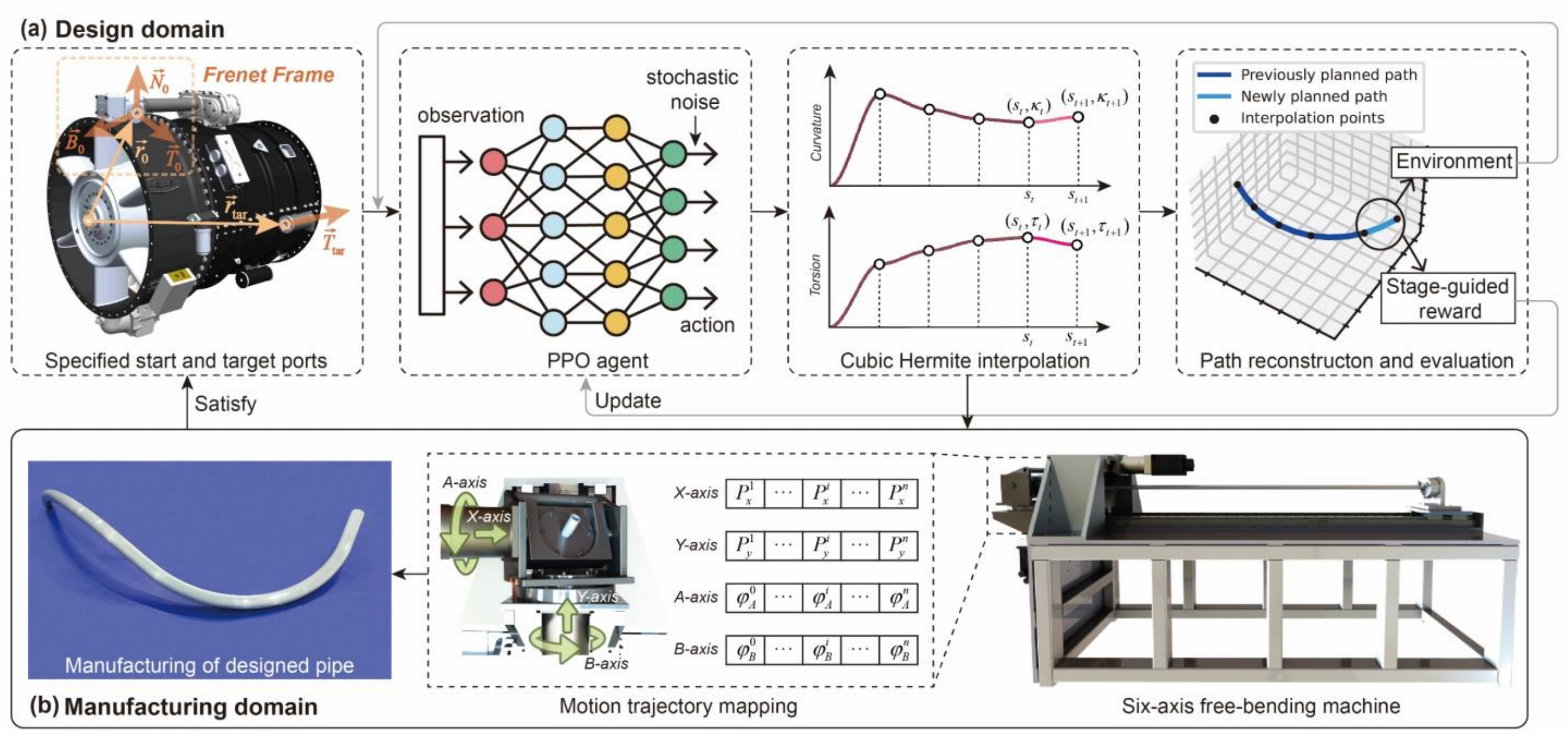

**Fig. 5.** Framework of Frenet-based pipe routing optimization.

## 4. Experiments and results comparison

This section presents the experimental validation of our proposed FPRO framework. We begin by detailing the experimental setup, including the layout environment, FPRO parameter configurations, and the specific pipe routing tasks designed for evaluation. Subsequently, we conduct a comparative analysis against Cartesian-based methods, emphasizing improvements in path smoothness and manufacturability achieved through the Frenet frame representation and cubic Hermite interpolation. The performance of our approach, which incorporates stochastic noise injection and stage-guided reward function, is then benchmarked against other RL baselines. Further ablation studies are performed to validate the efficacy of these strategic enhancements. Finally, we demonstrate the practical efficacy of the FPRO framework through real-world bending experiments on a six-axis free-bending machine.

### *4.1. Experimental setup*

To simulate realistic pipe routing scenarios, we developed a detailed aeroengine model using SolidWorks. The simulation environment is defined via a layout potential energy map, where all engine accessories are considered obstacle regions. Within this complex aeroengine model, we designed eight distinct pipe layout tasks, each with unique start and target ports. For all tasks, the pipe diameter was consistently set to 25 mm.

We carried out the routing experiments in a Python-based simulation environment. The FPRO network architecture, implemented in PyTorch, was trained on a cloud platform equipped with a 32-vCPU AMD EPYC 9654 processor. This setup provided substantial computational resources for the RL process. Key hyperparameters settings for the FPRO framework are summarized in Table 1. We determined the model parameters through both automatic parameter tuning tools [34] and subsequent manual adjustments. Considering the dimensions of the bending machine and the forming limits of the 6060-T6 aluminum alloy pipes, we defined the minimum bending radius, $R_{min}$, as 100 mm. Bending pipes with a radius below this minimum is typically infeasible or leads to severe defects.

**Table 1**

Summary of key hyperparameter configurations in FPRO.

| Parameter | Settings |
|---|---|
| Total training steps | 4000000 |
| Maximum steps per episode | 64 |
| Batch size | 64 |
| Hidden layers of neural networks | 2 |
| Neurons of each hidden layer | 128 |
| Optimizer type | Adam |
| Learning rate | $3 \times 10^{-4}$ |
| Reward discount factor $\gamma$ | 0.95 |
| Clip ratio | 0.15 |
| Noise amplitude $\alpha$ | 0.01 |
| $s_{max}$ | 20 mm |
| $[w_1, w_2, w_3, w_4]$ | [0.0875, 0.005, 2.5, 1] |
| $r_{bonus}$ | 10 |
| $[w_{align}, w_{shoot}]$ | [20, 50] |
| $[\varepsilon_{align}, \varepsilon_{shoot}, \varepsilon_{final}]$ | [200, 0.1, 0.05] |
| $R_{min}$ | 100 mm |

### *4.2. Performance comparison with Cartesian-based methods*

Cartesian-based pipe routing approaches commonly represent the pipe path using parametric spline curves, such as Non-Uniform Rational B-Splines (NURBS). To enhance path smoothness, these methods often employ high-order splines and incorporate penalty terms for curvature and torsion into the reward function. This section presents a comparative analysis between our proposed Frenet-based method, FPRO, and a representative Cartesian-based method, self-learning pipe routing (SLPR) [5]. Both FPRO and SLPR are RL-based routers that utilize PPO as their core planning algorithm. The original SLPR method generates paths using cubic NURBS curves with a maximum curvature constraint. To establish a more robust baseline, we also developed a modified version, *m*-SLPR, which employs quartic NURBS curves to ensure $C^3$ continuity and incorporates a minimum bending radius constraint to account for manufacturability.

Visualizations of the pipe layouts generated by FPRO, *m*-SLPR, and SLPR are presented in Fig. 6. Although all three methods successfully connect the start and target ports, their resulting paths exhibit distinct geometric characteristics. Specifically, SLPR tends to produce shorter paths but at the cost of smoothness, with its routes containing numerous twists and sharp turns. In contrast, *m*-SLPR utilizes quartic NURBS curves, which provide $C^3$ continuity and result in improved overall path smoothness. However, noticeable inflections persist in its routes for Pipe-3 and Pipe-4. Furthermore, the path for Pipe-4 generated by *m*-SLPR penetrates an obstacle. This failure likely stems from the agent's difficulty in simultaneously satisfying multiple conflicting constraints (e.g., path length, obstacle avoidance, curvature, and manufacturability). In this instance, the agent appears to have prioritized other constraints over collision avoidance. Ultimately, the paths generated by our FPRO method demonstrate superior smoothness and are entirely collision-free, delivering the most satisfactory and reliable overall performance.

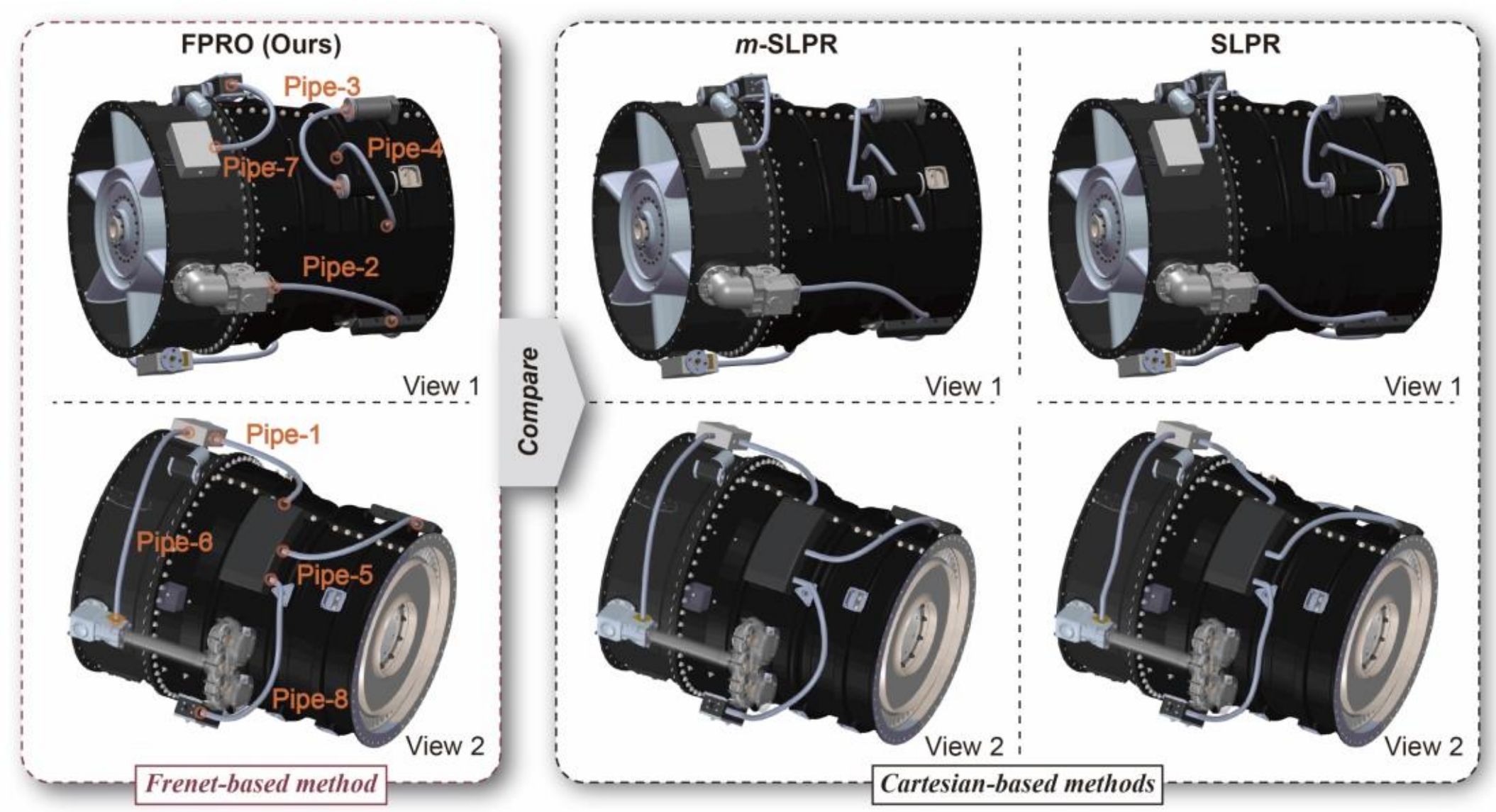

**Fig. 6.** Visualization of the layout schemes for FPRO and Cartesian-based methods.

To further investigate the geometric properties of the generated paths, we analyzed their curvature and torsion profiles, as depicted in Fig. 7 and Fig. 8, respectively. In these plots, the horizontal axis represents the arc length parameter, while the vertical axis denotes the magnitude of curvature or torsion. For both SLPR and *m*-SLPR, the curvature profiles exhibit sharp spikes near the path endpoints, which correspond directly to the visual inflection points observed in Fig. 6. The torsion profiles for SLPR show significant fluctuations across most tasks, with stable behavior observed only for Pipe-6 and Pipe-8. This instability is an inherent limitation of cubic NURBS curves, which cannot guarantee torsion continuity. Although the torsion profiles for *m*-SLPR are slightly more stable than those of SLPR, they still exhibit abrupt jumps, particularly for Pipe-1, Pipe-6, Pipe-7, and Pipe-8. These spikes and fluctuations in the geometric profiles can severely compromise the manufacturing process by causing erratic tool velocities and accelerations, which may lead to mechanical shock.

In stark contrast, the curvature and torsion profiles generated by FPRO remain remarkably stable, maintaining low magnitudes and exhibiting no abrupt changes. This advantage stems directly from our path generation strategy, which models the curvature and torsion profiles using cubic Hermite interpolation. For the current scenarios, we set the allowable ranges for curvature and torsion in FPRO to [0, 0.01] and [-0.005, 0.005], respectively. This approach inherently ensures path continuity and stability, yielding exceptionally smooth paths that are conducive to high-quality manufacturing and improved kinematic performance of the bending die.

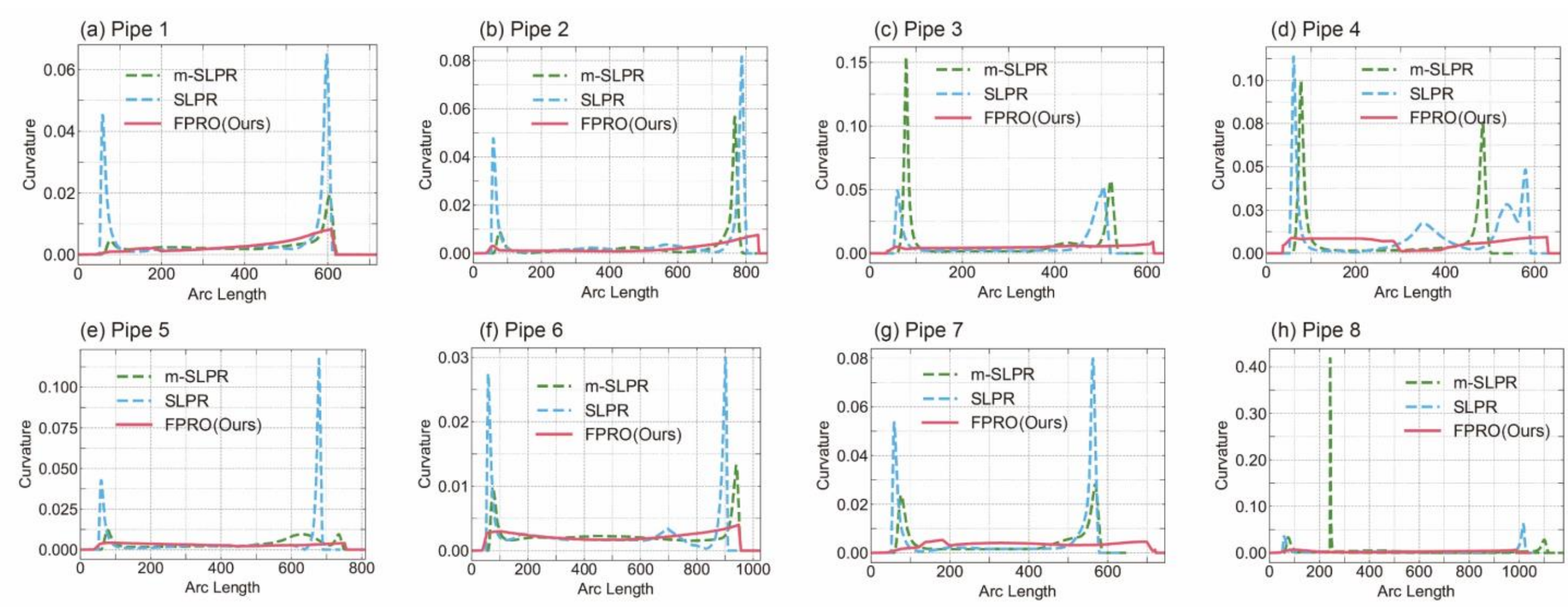

**Fig. 7.** Curvature comparison between FPRO and Cartesian-based methods.

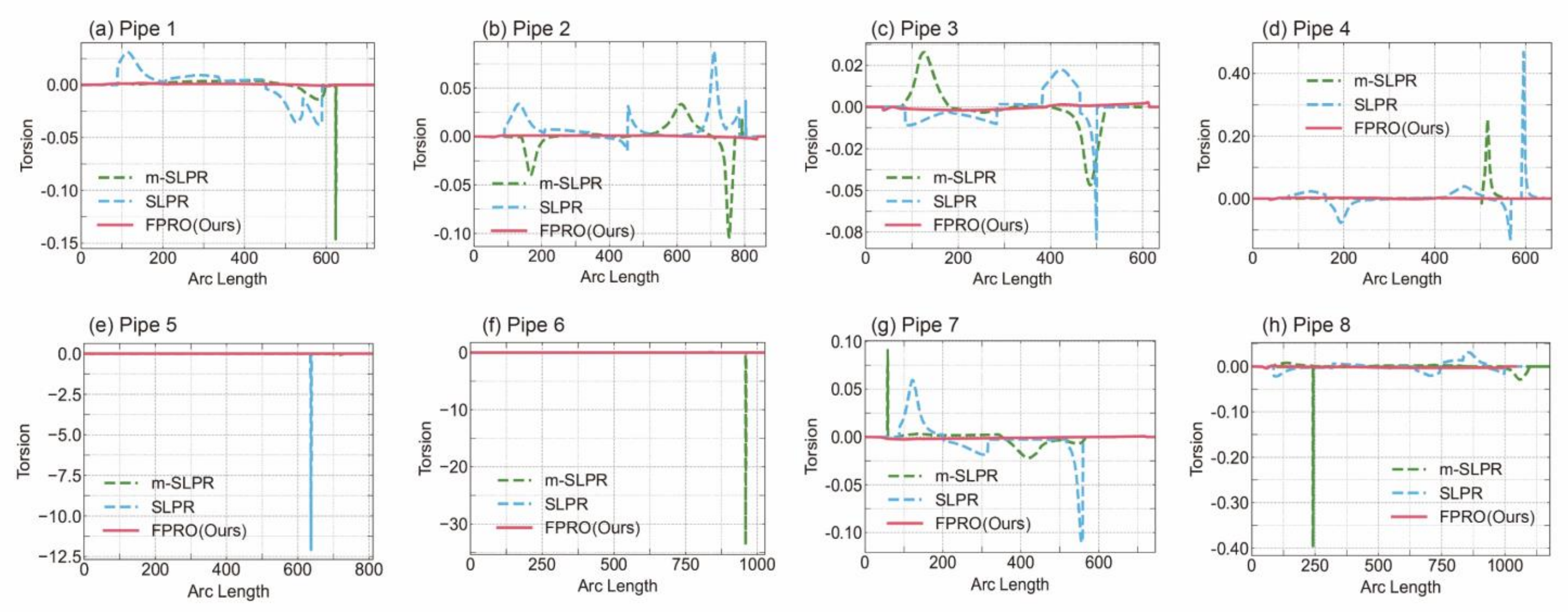

**Fig. 8.** Torsion comparison between FPRO and Cartesian-based methods.

Finally, we present a quantitative evaluation of FPRO, *m*-SLPR, and SLPR. Performance is assessed based on path length, obstacle avoidance, and manufacturability, using three metrics: pipe length (PL), a collision-free indicator (CFI), and the manufacturability violation rate (MVR). The MVR is calculated as the percentage of points sampled along the pipe path that violate the minimum bending radius constraint. Detailed results for all tasks are compiled in Table 2, with the best results marked in bold. The results show that SLPR achieves the shortest average path length, followed by *m*-SLPR and FPRO. However, SLPR's shorter routes come at a significant cost to physical feasibility, evidenced by an average MVR of 59.1%. This indicates that over half of the generated pipe sections would be unrealizable in a real-world manufacturing process. By incorporating the minimum bending radius constraint, *m*-SLPR improves upon this, reducing the average MVR to 31.0%. This improvement, however, compromises its obstacle avoidance capability, resulting in a collision in the Pipe-4 task. In contrast, our proposed FPRO method excels across all metrics. It not only achieves a 100% collision-free success rate across all tasks but also reduces the MVR to 0% in every case. This perfect manufacturability is a direct benefit of our Frenet-based framework, which inherently prevents the generation of non-manufactured solutions at the source. Unlike Cartesian-based approaches that treat manufacturability as a penalty to be minimized, our method adopts a DFM paradigm from the outset. By guaranteeing manufacturable and collision-free routes, FPRO demonstrates superior practical value and engineering significance.

**Table 2**

Layout performance comparison between FPRO and Cartesian-based methods.

| Pipe index | FPRO (Ours) | | | *m*-SLPR | | | SLPR | | |
|---|---|---|---|---|---|---|---|---|---|
| | PL (mm) | CFI | MVR (%) | PL | CFI | MVR (%) | PL | CFI | MVR (%) |
| Pipe-1 | 717.46 | ✓ | **0** | 681.05 | ✓ | 22.2 | **662.24** | ✓ | 69.0 |
| Pipe-2 | 861.76 | ✓ | **0** | **849.98** | ✓ | 33.2 | 854.05 | ✓ | 57.8 |
| Pipe-3 | 634.76 | ✓ | **0** | 598.66 | ✓ | 41.8 | **571.63** | ✓ | 58.6 |
| Pipe-4 | 656.37 | ✓ | **0** | **561.45** | × | 40.0 | 642.15 | ✓ | 77.2 |
| Pipe-5 | 805.43 | ✓ | **0** | 810.49 | ✓ | 26.0 | **740.33** | ✓ | 49.0 |
| Pipe-6 | 1025.77 | ✓ | **0** | 1016.48 | ✓ | 19.8 | **960.87** | ✓ | 58.8 |
| Pipe-7 | 742.25 | ✓ | **0** | 647.02 | ✓ | 35.8 | **629.63** | ✓ | 50.0 |
| Pipe-8 | **1039.90** | ✓ | **0** | 1177.39 | ✓ | 29.0 | 1076.85 | ✓ | 52.6 |
| Average | 810.46 | **0/8** | **0** | 792.82 | 1/8 | 31.0 | **767.22** | **0/8** | 59.1 |

*4.3. Performance comparison with RL baselines*

To evaluate the performance of our proposed FPRO framework in layout optimization, we conducted a comparative analysis against several state-of-the-art RL algorithms and performed ablation studies on the core components of FPRO. The baseline algorithms included soft actor-critic (SAC) [35], twin delayed deep deterministic policy gradient (TD3) [36], and phasic policy gradient (PPG) [37]. To ensure a fair comparison, all baselines were integrated into our Frenet-based routing framework, utilizing the same observation space, action space, and reward function as FPRO. The implementations were based on the robust CleanRL library [38].

To examine the contributions of our strategic enhancements, we evaluated two FPRO variants: FPRO-N, which excludes the stochastic noise injection module, and FPRO-R, which removes the stage-guided reward mechanism. For all experiments, we saved the policy that yielded the best performance during the training for the final evaluation. We assessed performance using the PL, CFI, and MVR metrics, as defined in Section 4.2. Furthermore, we introduce the alignment loss $l_{align}$, defined in Eq. (32), to quantitatively measure the alignment between the pipe's terminal segment and the target port, where a smaller value indicates a better result.

The comparative results are presented in Table 3, with the best performance for each metric highlighted in bold. These results demonstrate the superior performance of FPRO. Compared to state-of-the-art baselines, FPRO significantly outperforms TD3, SAC, and PPG across all evaluation metrics. Notably, although TD3 and PPG successfully avoided obstacles in all tasks, they struggled with terminal alignment, yielding $l_{align}$ values approximately five to six times higher than those of FPRO. SAC exhibited the poorest performance, encountering collisions in three of the eight tasks and recording the highest MVR. The ablation study reveals that removing either the stage-guided reward (FPRO-R) or the noise injection (FPRO-N) results in a tangible degradation in performance compared to the complete FPRO model. Although both variants achieved perfect obstacle avoidance and manufacturability rates, demonstrating the robustness of the underlying framework, their alignment precision and path efficiency were inferior. Specifically, the complete FPRO model reduced the $l_{align}$ by approximately 50% compared to its ablated versions and generated shorter paths. These findings confirm that both the stage-guided reward and the noise-enhanced exploration are crucial components that synergistically contribute to the final, high-performance policy.

**Table 3**

Performance comparison between FPRO and RL baselines.

| Methods | $l_{align}$ | PL (mm) | CFI | MVR |
|---|---|---|---|---|
| TD3 | 0.476 | 935.57 | **0** | 0.50% |
| SAC | 0.617 | 874.36 | 3/8 | 2.09% |
| PPG | 0.397 | 849.75 | **0** | 0.14% |
| FPRO-N | 0.140 | 834.92 | **0** | **0** |
| FPRO-R | 0.165 | 834.91 | **0** | **0** |
| FPRO (Ours) | **0.084** | **810.46** | **0** | **0** |

To further evaluate the learning efficiency and peak performance of the algorithms, we analyzed their average returns during training and the maximum return achieved on each task, as illustrated in Fig. 9. Fig. 9 (a) presents the learning curves of the average returns. It clearly shows that FPRO exhibits superior learning efficiency and faster convergence, with its return curve rising more rapidly and stabilizing at a significantly higher level than those of all other methods. The ablation study provides further insights: although FPRO-N maintains stable learning behavior, its final average return is lower than that of the complete FPRO model. This finding highlights the effectiveness of the noise injection strategy in enabling the agent to escape local optima and discover better policies. In contrast, the baseline algorithms lag considerably. PPG's learning curve is unstable, while TD3 and SAC struggle to learn effectively and fail to converge to satisfactory policies. Fig. 9 (b) displays the maximum return achieved across the eight pipe routing tasks, reflecting each algorithm's ability to find high-quality solutions. FPRO achieves the highest return on half of the tasks, including Pipe-3, 4, 5, and 8, which demonstrates its robust performance across diverse scenarios. Notably, FPRO-N also proves to be highly competitive, achieving slightly higher peak returns on Pipe 1, 2, and 6, while PPG yields the best performance on Pipe 7. In summary, FPRO not only converges more reliably to high-quality policies but also frequently discovers peak-performing solutions, highlighting the overall superiority of our proposed framework.

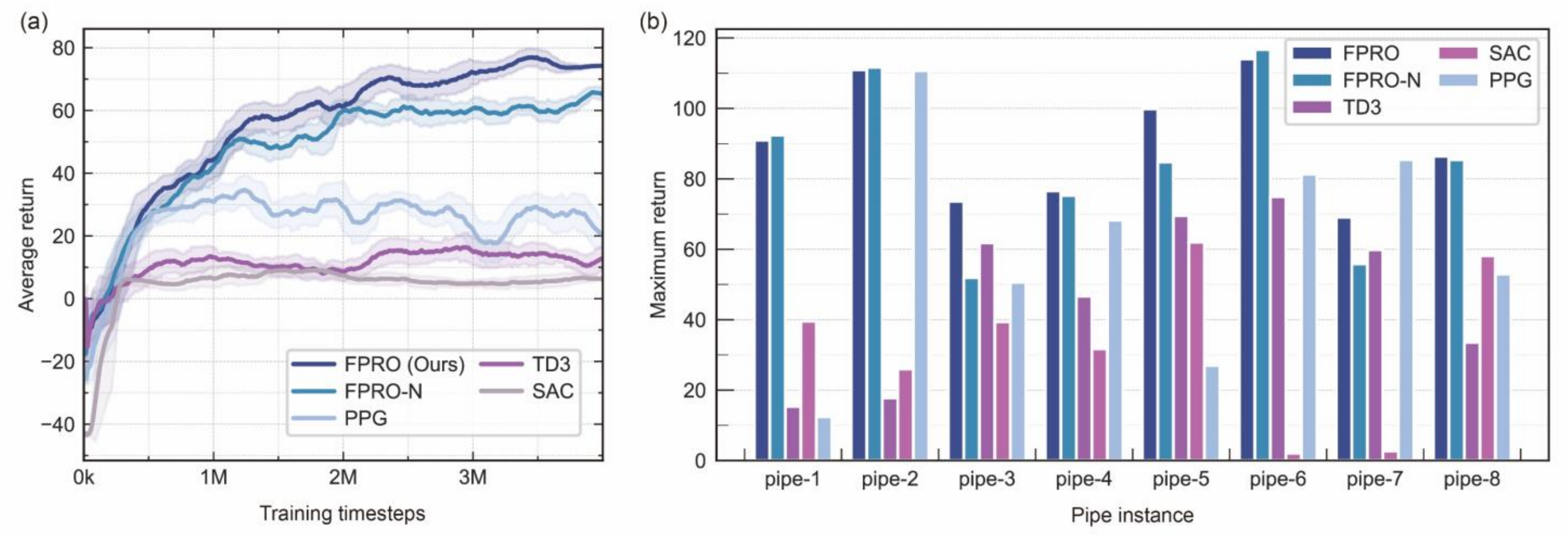

**Fig. 9.** Average and maximum return comparison between FPRO and RL baselines.

*4.4. Real-world validation on a six-axis free-bending machine*

To experimentally validate the manufacturability of pipes designed using the FPRO framework, a real-world bending trial was conducted on a self-developed six-axis free-bending machine. Pipe-6 was selected as the representative case due to its considerable length and complex spatial geometry, making it a rigorous test for the practical feasibility of the proposed methodology. The procedure began by translating the digital design into machine-executable instructions. Using the unified mapping formula defined in Eq. (2), the curvature and torsion profiles of Pipe-6 were converted into motion trajectories for each axis of the bending die, as illustrated in Fig. 10 (a). A constant feed rate (Z-axis velocity) of 1.5 mm/s was employed, and a correction factor of $k$ = 1.5, determined through empirical tuning, was applied to compensate for material springback. These trajectories were then uploaded to the free-bending machine, as shown in Fig. 10 (b), to bend the pipe made of 6061-T6 aluminum alloy. The final manufactured free-form pipe and its point clouds scanning process are depicted in Fig. 11. To quantitatively assess manufacturing accuracy, a handheld laser 3D scanner (FreeScan Combo+) was used to capture a high-resolution point cloud of the manufactured pipe, which served as the basis for geometric comparison with the original digital design.

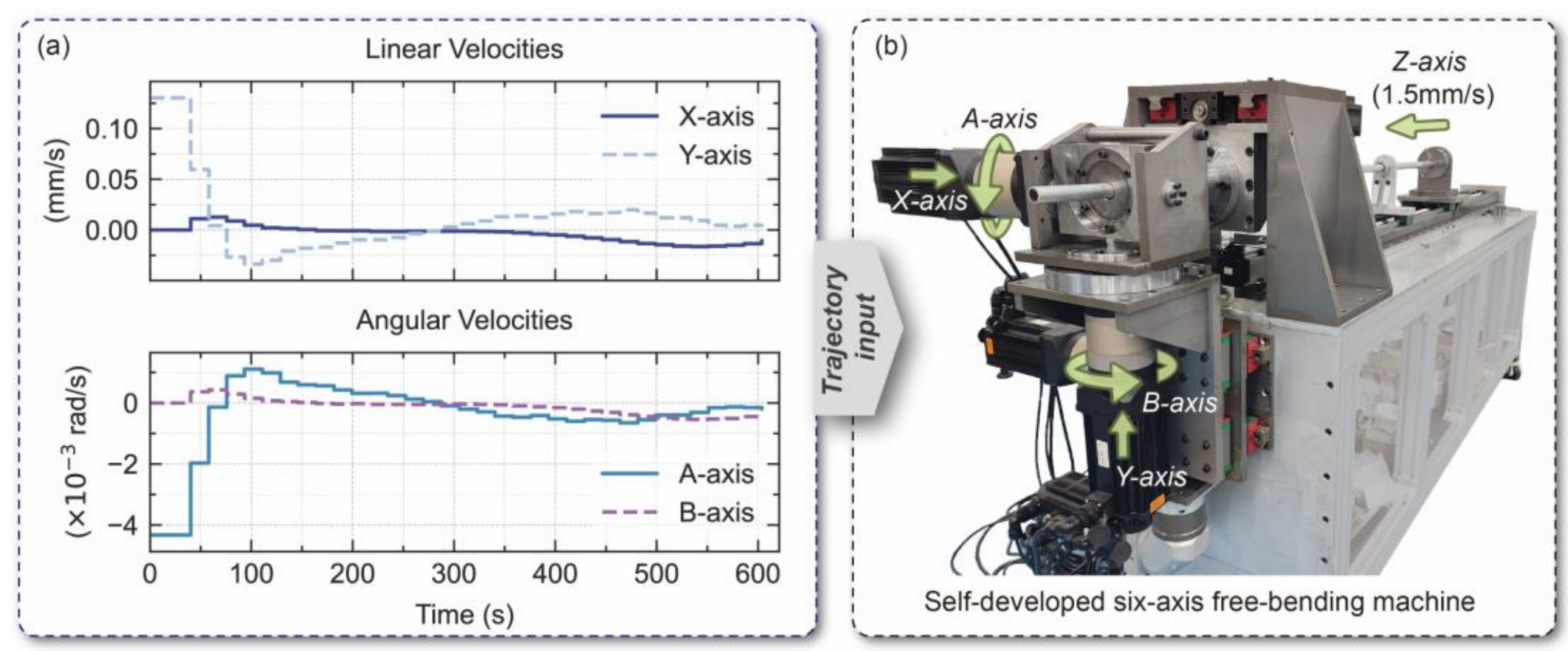

**Fig. 10.** Real-world bending experiments for the designed free-form pipe. (a) Velocity trajectories of each axis. (b) Self-developed six-axis free-bending machine.

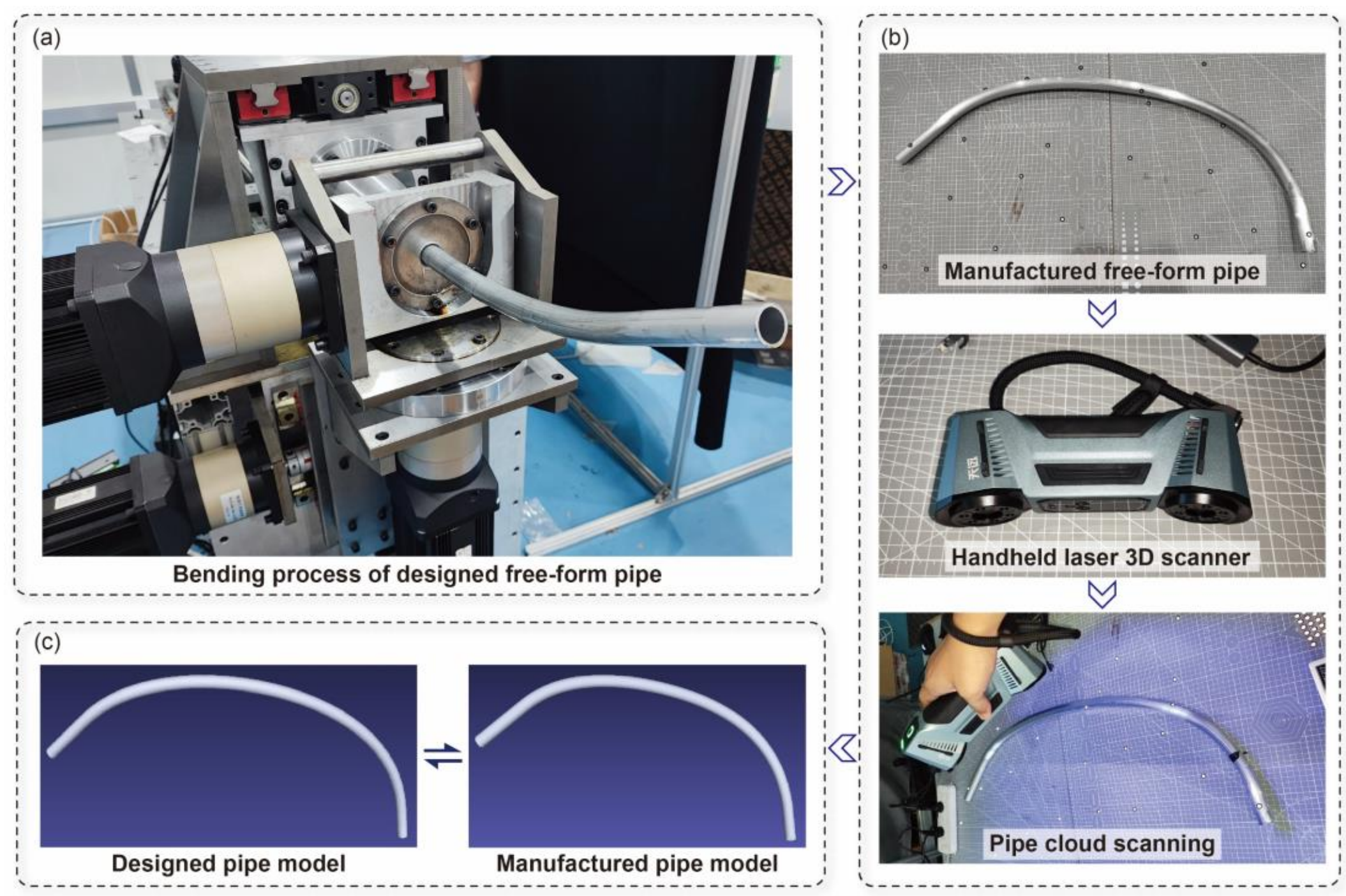

**Fig. 11.** Manufactured free-form pipe and point cloud acquisition process. (a) Bending process of designed free-form pipe. (b) Pipe cloud scanning of the manufactured pipe. (c) Geometric comparison between the manufactured and designed pipe models.

Fig. 11 (c) reveals a high degree of visual alignment between the manufactured pipe and the original digital model. To assess the geometric fidelity in detail, we extracted the central axis of the manufactured pipe from the point cloud data using an iterative detection sphere method [39]. Fig. 12 presents a comparison between the extracted experimental axis and the designed axis. In this figure, the local deviation is defined as the minimum Euclidean distance from each point on the experimental axis to the design curve. These deviations are visualized using a color map, where blue denotes minimal deviation, and red indicates larger discrepancies. The two axes exhibit substantial overlap, particularly along the main body of the pipe, indicating excellent geometric conformity. However, deviations increase near the start and end sections, reaching a maximum of 39.21 mm. This behavior corresponds to the inherent characteristics of the free-bending process, which requires straight lead-in and lead-out segments to initiate and complete the bending operation. These segments, known as "transition sections," deviate from the original design and are typically treated as manufacturing allowances that are trimmed in practical applications.

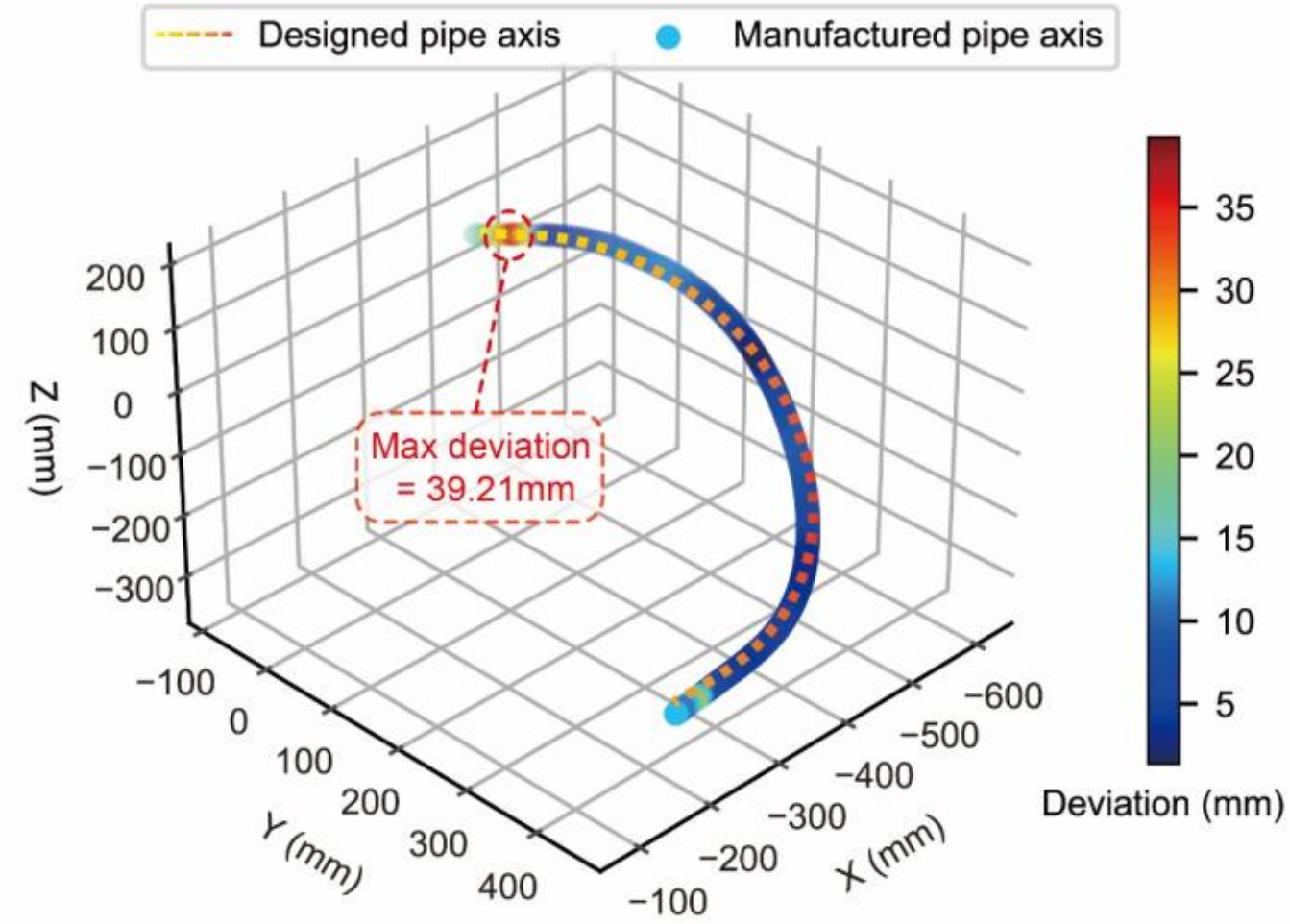

**Fig. 12.** Visualization of the spatial deviation distribution between the designed and manufactured pipe axes.

To provide a quantitative evaluation, we employed several metrics to assess the similarity between the experimental and designed pipe axes, including the longest common subsequence (LCSS), Fréchet distance (FD), dynamic time warping (DTW), and edit distance (ED) [40]. LCSS evaluates structural similarity by identifying the longest matching subsequence of points whose spatial distance falls within a specified tolerance. For this experiment, the threshold was set to 25 mm, corresponding to the diameter of the pipe. FD measures global similarity by computing the minimum "leash" length needed to traverse both curves synchronously. DTW determines the optimal alignment by non-linearly warping the time axis and calculating the minimum cumulative distance under this alignment. ED quantifies dissimilarity by computing the minimum number of operations (insertions, deletions, or substitutions) required to transform one trajectory into the other.

The quantitative similarity results, presented in Table 4, strongly support the visual observations from Fig. 12. For the complete pipe, the LCSS ratio reaches 0.98, indicating that 98% of the pipe axis points lie within the spatial tolerance of the designed path, confirming the overall geometric accuracy. However, relatively high values for the Fréchet distance (FD), dynamic time warping (DTW), and edit distance (ED) suggest notable local deviations, alignment costs, and structural differences, primarily attributed to the transition sections. A substantial improvement is observed when evaluating only the core section of the pipe, with the LCSS ratio rising to a perfect 1.00, indicating complete structural consistency. Notably, the FD drops by 70% to 11.79, confirming that the most significant deviations are concentrated in the transition zones. Concurrently, DTW decreases by 26%, and ED is reduced from 17 to 12, reflecting lower alignment cost and enhanced structural fidelity. These findings validate that the FPRO framework can produce physically manufacturable pipe paths that closely replicate their digital design, underscoring its strong practical applicability and engineering value.

**Table 4**

Quantitative similarity assessment between the experimental and designed pipe axis.

| Pipe Version | LCSS | FD | DTW | ED |
|---|---|---|---|---|
| Complete pipe (with transitions) | 0.98 | 39.21 | 1642.30 | 17 |
| Core pipe (transitions removed) | 1.00 | 11.79 | 1209.72 | 12 |

FPRO adopts a DFM paradigm by enforcing continuity of curvature and torsion and adhering to the minimum bending radius constraint. Although these measures help reduce springback during forming, the phenomenon cannot be eliminated because it is governed by material mechanics. Springback, together with the inevitable lead-in and lead-out transition sections, remains a principal source of deviation between the manufactured pipe and its digital design. Recent studies have demonstrated that graph neural networks (GNNs) can model the pipe bending process with high fidelity and accurately predict the post-springback axis [12, 41], suggesting a promising direction for extending the FPRO framework. The path generated by FPRO could be fed into a GNN to estimate its final shape after bending and springback. The predicted deviation from the target design would then be quantified and incorporated as an additional penalty term in FPRO's reward function. In this way, the optimization objective evolves from achieving purely geometric optimality to proactively compensating for manufacturing deviations, thereby establishing a closed-loop optimization framework that seamlessly integrates digital design with predictive physical modeling.

## 5. Conclusion

This paper presents FPRO, an integrated reinforcement learning framework that directly generates manufacturable free-form pipe paths for complex aeroengine environments. To address the highly coupled and nonlinear manufacturing constraints inherent in the Cartesian space, FPRO reformulates the routing task as a BVP within the Frenet frame. In this paradigm, curvature and torsion profiles are generated using cubic Hermite interpolation, and the corresponding spatial path is reconstructed via the Frenet–Serret formulas. The selection of interpolation points is modeled as a sequential decision-making problem and solved using the PPO algorithm, which incorporates stochastic noise to encourage exploration and a stage-guided reward function to enhance robustness. The resulting pipe path can be directly converted into the motion trajectory of the bending die using a unified mapping formula, enabling practical manufacturing. The effectiveness of the proposed FPRO framework is validated through comprehensive experiments, with the primary findings summarized as follows:

(1) The Frenet-based approach demonstrated clear superiority over traditional Cartesian-based methods, including SLPR and m-SLPR. Cartesian methods struggled to balance conflicting constraints, resulting in either collisions or high rates of manufacturability violations. In contrast, the FPRO consistently generated collision-free and fully manufacturable paths across all routing tasks, featuring smooth and stable geometric profiles.

(2) Comparative experiments with state-of-the-art RL baselines show that FPRO achieves superior learning efficiency and faster convergence than established algorithms like TD3, SAC, and PPG. FPRO produced routing solutions with the best overall performance in terminal alignment accuracy, path length, obstacle avoidance, and manufacturability. Additionally, comprehensive ablation studies confirmed that both the stage-guided reward function and the noise injection mechanism are essential, working synergistically to achieve high-performance policies.

(3) The framework's practical viability and manufacturing fidelity were validated through a real-world bending experiment. A pipe designed using FPRO was successfully formed on a six-axis free-bending machine. Quantitative analysis of the 3D-scanned physical component revealed an excellent geometric agreement with the digital model, confirming that FPRO effectively bridges the gap between digital design and physical manufacturing.

In summary, this research introduces a comprehensive and practical DFM paradigm for free-form pipe routing. By embedding manufacturability constraints into the design process, FPRO serves as a robust framework for generating high-quality, reliable, and physically realizable pipe layouts. Although FPRO ensures manufacturable designs, inherent forming defects in the bending process may still lead to deviations between the manufactured and digital pipes. Future work will focus on integrating FPRO with a surrogate model that can predict these defects in real time. This enhancement will enable the agent to proactively compensate for manufacturing deviations, thereby establishing a closed-loop optimization system that seamlessly links digital design with predictive physical modeling.

## Acknowledgements

This paper is funded by the National Natural Science Foundation of China (52275274), the Joint Funds of the National Natural Science Foundation of China (U20A20287), the Pioneer and Leading Goose R&D Program of Zhejiang, China (2024C01197), the Public Welfare Technology Application Projects of Zhejiang Province, China (LGG22E050008), the State Key Laboratory of Intelligent Manufacturing Equipment and Technology Open Project, China (IMETKF2024007). Zhejiang Province leading innovative entrepreneurial team project (2022R01012). The Shandong Offshore Engineering Facility & Material Innovation Entrepreneurship Community (SOFM-IEC), China (GTP-2501).

## CRediT authorship contribution statement

**Caicheng Wang:** Conceptualization, Methodology, Software, Writing - Original, Writing - Review & Editing, Dara Curation, Validation, Visualization. **Zili Wang:** Formal analysis, Writing - Review & Editing, Funding acquisition. **Shuyou Zhang:** Project administration, Funding acquisition. **Yongzhe Xiang:** Investigation, Software. **Zheyi Li:** Investigation, Visualization. **Liangyou Li**: Project administration, Funding acquisition. **Jianrong Tan:** Project administration, Funding acquisition.

## Competing Interests

Authors declare that they have no conflict of interest.

## Data and code availability

Data will be made available on request.